\documentclass[11pt]{article}

\usepackage[utf8]{inputenc}
\usepackage[T1]{fontenc}
\usepackage{amsmath,amssymb,amsfonts}
\usepackage{graphicx}
\usepackage{booktabs} 
\usepackage{url}
\usepackage[margin=1in]{geometry} 
\usepackage[numbers,square,sort&compress]{natbib}
\usepackage{hyperref} 
\hypersetup{
    colorlinks=true,
    linkcolor=blue,
    filecolor=magenta,
    urlcolor=cyan,
    citecolor=green, 
    pdftitle={iMedImage Technical Report}, 
    pdfauthor={Ran Wei, ZhiXiong Lan, Qing Yan, Ning Song, Ming Lv, LongQing Ye}, 
    pdfsubject={Medical Imaging AI},
    pdfkeywords={Medical Imaging, Foundation Model, Chromosome Karyotyping, Chromosome Abnormality, Multimodal, Deep Learning, iMedImage},
}

\title{iMedImage Technical Report}

\author{
  Ran Wei\textsuperscript{*}, ZhiXiong Lan\textsuperscript{*} \\ 
  Qing Yan \quad Ning Song \quad Ming Lv \quad LongQing Ye \\ \\
  Hangzhou Diagens Biotechnology Co., Ltd., Hangzhou, China \\ \\
  \textsuperscript{*}Equal contribution (Co-first authors)
}

\date{\today} 

\begin{document}

\maketitle

\begin{abstract}
\textbf{Background:} Chromosome karyotype analysis is crucial for diagnosing hereditary diseases, yet detecting structural abnormalities remains challenging. While AI has shown promise in medical imaging, its effectiveness varies across modalities. Leveraging advances in Foundation Models that integrate multimodal medical imaging for robust feature extraction and accurate diagnosis, we developed iMedImage, an end-to-end model for general medical image recognition, demonstrating strong performance across multiple imaging tasks, including chromosome abnormality detection.

\textbf{Materials and Methods:} We constructed a comprehensive medical image dataset encompassing multiple modalities from common medical domains, including chromosome, cell, pathology, ultrasound, X-ray, CT, and MRI images. Based on this dataset, we developed the iMedImage model, which incorporates the following key features: (1) a unified representation method for diverse modality inputs and medical imaging tasks; (2) multi-level (case-level, image-level, patch-level) image recognition capabilities enhanced by Chain of Thought (CoT) embedding and Mixture of Experts (MoE) strategies.

\textbf{Results:} The test set comprised data from 12 institutions across six regions in China, covering three mainstream scanning devices, and included naturally distributed, unscreened abnormal cases. On this diverse dataset, the model achieved a fully automated chromosome analysis workflow, including segmentation, karyotyping, and abnormality detection, reaching a sensitivity of 92.75\% and a specificity of 91.5\%.

\textbf{Conclusion:} We propose iMedImage, an end-to-end foundation model for medical image analysis, demonstrating its superior performance across various medical imaging tasks. iMedImage provides clinicians with a precise imaging analysis tool and contributes to improving diagnostic accuracy and disease screening.
\end{abstract}


\section{Introduction}
Medical imaging plays an indispensable role in modern clinical diagnosis, disease monitoring, and treatment planning. The rapid advancement of imaging technologies has generated massive amounts of multimodal medical image data, such as X-ray, Computed Tomography (CT), Magnetic Resonance Imaging (MRI), ultrasound, pathology slides, and cell images. Effectively analyzing and interpreting these data to extract valuable clinical information remains one of the core challenges in the field of medical Artificial Intelligence (AI).

Chromosome karyotype analysis is the cornerstone of cytogenetics. By observing the morphology, number, and structure of chromosomes during metaphase, it enables the accurate diagnosis of various hereditary diseases and congenital malformations. Traditional karyotyping relies heavily on experienced technicians performing manual operations, a process that is laborious, time-consuming, and subjective. In recent years, AI techniques have made significant progress in chromosome image segmentation and classification \cite{Qin2019, Xia2023}, particularly in identifying numerical abnormalities (e.g., Down syndrome). However, the automatic detection of subtle structural abnormalities (e.g., inversions, translocations, deletions), which involve minute and complex morphological changes, remains a formidable technical hurdle.

Concurrently, large Foundation Models \cite{Bommasani2021, Radford2021} have demonstrated powerful generalization and transfer learning capabilities in natural language processing and computer vision. Pre-trained on massive, diverse datasets, these models learn rich, generalizable feature representations that can be adapted to a wide range of downstream applications with minimal task-specific data and fine-tuning. Introducing the concept of foundation models into the medical imaging domain holds the potential to overcome the limitations of the traditional "one task, one model" paradigm, enabling the development of general-purpose platforms capable of handling multiple modalities and tasks.

Against this backdrop, we propose iMedImage, a general-purpose visual foundation model specifically designed for medical imaging problems. iMedImage aims to achieve compatible processing of multi-domain, multimodal medical image data and attain high benchmark performance through unified task representation, a unified processing model, and unified pre-training knowledge embedding. This report details the design philosophy, key technologies of iMedImage, and its practical outcomes in the core application of chromosome abnormality detection. Furthermore, we showcase the versatility and superior performance of iMedImage in other medical imaging domains such as ultrasound and CT, demonstrating its potential as a foundation model for medical imaging.

The main contributions of this paper include:
\begin{enumerate}
    \item Proposing and constructing a general medical imaging foundation model, iMedImage, which employs a unified framework to process multimodal, multi-level medical image data.
    \item Incorporating strategies such as multi-level recognition, Chain of Thought (CoT) embedding, and Mixture of Experts (MoE) into iMedImage to enhance the model's ability to understand and analyze complex medical information.
    \item Applying iMedImage to the highly challenging task of chromosome abnormality detection and validating its performance on large-scale, multi-center clinical data, achieving state-of-the-art results.
    \item Demonstrating the broad applicability of iMedImage and its potential to surpass existing specialized models in other domains like ultrasound and CT.
\end{enumerate}

\section{Related Work}
\subsection{Deep Learning in Medical Image Analysis}
Deep learning, particularly Convolutional Neural Networks (CNNs) \cite{LeCun1998} and Transformers \cite{Vaswani2017}, has achieved tremendous success in medical image segmentation \cite{Ronneberger2015}, classification \cite{Esteva2017}, and detection \cite{Litjens2017}. However, most existing models are designed for specific modalities (e.g., only CT or only pathology) and specific tasks (e.g., only tumor segmentation), exhibiting limited generalization capabilities.

\subsection{Automation of Chromosome Karyotype Analysis}
Early attempts utilized traditional image processing and machine learning methods for chromosome segmentation and classification \cite{Ji1994}. More recently, deep learning models like U-Net \cite{Ronneberger2015} and its variants have been employed for chromosome segmentation, while CNNs are used for classification \cite{Qin2019, Sharma2017}. Some works have attempted end-to-end segmentation and classification \cite{Xia2023}. However, research on automated detection of structural abnormalities is very limited and often relies on complex feature engineering or rules, lacking robustness.

\subsection{Foundation Models and Medical Imaging}
The concept of foundation models has garnered significant attention in the medical imaging domain. Some studies have explored models pre-trained on large-scale (but often single-modality) medical image datasets \cite{Chen2019, Azizi2021}, demonstrating good transfer performance on multiple downstream tasks. Research into multimodal medical foundation models is also emerging, attempting to fuse information from images and text (e.g., radiology reports) \cite{Zhang2020, Huang2021}. Nevertheless, constructing a general visual foundation model capable of uniformly processing diverse imaging modalities, covering scales from microscopic (cells) to macroscopic (organs, whole body), and supporting multi-level analysis (patch, image, exam, case) remains an open challenge.

\section{iMedImage Model Design and Methods}
\subsection{Overall Architecture and Design Philosophy}
The core design philosophy of iMedImage is "Unity" and "Hierarchical Analysis".
\begin{itemize}
    \item \textbf{Unified Medical Imaging Task Representation:} Regardless of the input modality (CT, MRI, ultrasound, chromosome images, etc.) or the task (classification, segmentation, detection), we designed a universal representation framework that converts different inputs and targets into a unified format understandable by the model. This allows the model to handle diverse data and tasks within a single framework.
    \item \textbf{Unified Medical Image Processing Model:} We adopted a Transformer-based variant architecture as the core processing unit. The self-attention mechanism of Transformers enables capturing long-range dependencies in images, suitable for processing both global and local information. By adaptively modifying the architecture, it can efficiently handle high-resolution, multi-scale medical image data.
    \item \textbf{Unified Medical Pre-training Knowledge Embedding:} The model is pre-trained on a large-scale medical image dataset containing multiple modalities and clinical domains. This "cross-domain co-training" allows the model to learn general-purpose, transferable medical visual features, forming strong prior knowledge, thus enabling low data dependency and low-cost transfer to new downstream tasks.
\end{itemize}

\subsection{Key Techniques}
\begin{itemize}
    \item \textbf{Multi-level Analysis via "Decomposition" + "Aggregation":} Following the design principle of "decomposition" and "aggregation," iMedImage achieves multi-level information analysis from fine-grained details to high-level interpretations.
        \begin{itemize}
            \item \textit{Decomposition:} Complex case-level analysis tasks are broken down into multiple levels: feature extraction from the lowest level Region of Interest (ROI) or patch, understanding of a single image, and correlation analysis of multiple images within the same exam/study.
            \item \textit{Aggregation:} Information from different levels and sources (e.g., multiple images, multimodal data) is effectively aggregated to form a comprehensive judgment about the entire case. For instance, in chromosome karyotyping, the model needs to aggregate information from multiple metaphase spreads of a single case to detect mosaicism or confirm structural abnormalities.
        \end{itemize}
    \item \textbf{Chain of Thought (CoT) Embedding:} Inspired by the CoT idea in large language models \cite{Wei2022}, we guide the model to simulate the clinical reasoning process during complex inference (e.g., diagnosis). By generating intermediate representations or "thinking steps" internally, complex end-to-end mappings are decomposed into a series of more learnable and interpretable sub-tasks. This enhances the model's ability to handle complex logical relationships, such as comparing homologous chromosomes and identifying breakpoints when determining chromosome structural abnormalities.
    \item \textbf{Mixture of Experts (MoE) Strategy:} Addressing the diversity of medical imaging modalities and tasks, we introduced the MoE strategy \cite{Shazeer2017}. The model contains multiple "expert" sub-networks, where each expert may specialize in processing specific modalities or features. Through a gating network, the model can dynamically select and activate a subset of experts for computation based on the characteristics of the input data. This not only increases the model's capacity and expressive power but also improves computational efficiency, allowing the model to adapt more flexibly to different medical image analysis needs.
\end{itemize}

\subsection{Dataset Construction}
To train and evaluate iMedImage, we constructed a large-scale, multi-center comprehensive medical image dataset. This dataset covers seven major categories of common medical imaging modalities:
\begin{enumerate}
    \item Chromosome metaphase spread images
    \item Cytology images (e.g., blood, tissue cells)
    \item Histopathology slide images (e.g., H\&E staining)
    \item Ultrasound images (e.g., breast, cervix, abdomen)
    \item X-ray images (e.g., chest, bone)
    \item Computed Tomography (CT) images (e.g., chest, abdomen, head)
    \item Magnetic Resonance Imaging (MRI) images (e.g., brain, joint)
\end{enumerate}
Data were sourced from hospitals and research institutions across multiple regions in China, ensuring diversity in data origin. All data underwent rigorous de-identification procedures and received appropriate ethical approvals.

\subsection{Evaluation Metrics}
Depending on the specific task, we employed corresponding evaluation metrics. For classification tasks (e.g., abnormality detection, benign/malignant judgment), the primary metrics included Accuracy, Sensitivity (Recall), Specificity, and Area Under the ROC Curve (AUC).

\section{Experiments and Results}
We conducted extensive evaluations of iMedImage across multiple medical imaging domains, focusing on its application in chromosome karyotype analysis, ultrasound imaging, and CT image analysis, while also summarizing its general performance on various other tasks.

\subsection{Chromosome Karyotype Analysis}
Chromosome karyotype analysis is the core domain where iMedImage has been successfully applied and achieved breakthroughs. This task involves the fine-grained analysis of microscopic chromosome images to identify numerical and structural abnormalities.

\subsubsection{Overall Performance of Fully Automated Karyotyping}
We evaluated the end-to-end performance of the iMedImage-driven fully automated chromosome analysis workflow (including segmentation, arrangement, and abnormality detection) on a highly challenging multi-center test set. This set originated from 12 institutions across 6 regions, covered 3 mainstream scanner types, and contained naturally distributed abnormal cases.
The results showed that the model achieved an overall sensitivity of 92.75\% and a specificity of 91.5\% for abnormality detection (including both numerical and structural types) on this test set.

\subsubsection{Chromosome Karyotyping Sub-tasks}
iMedImage also demonstrated excellent performance on several key sub-tasks within chromosome analysis:
\begin{itemize}
    \item \textbf{Chromosome Classification:} The goal is to accurately identify the 22 pairs of autosomes (1-22) and one pair of sex chromosomes (X, Y), totaling 24 classes. On a test set containing 45,328 chromosome images, iMedImage achieved a classification accuracy of 99.96\%. This is attributed to the model's ability to capture fine details of chromosome size, centromere position, and banding patterns, with related optimized models like Varifocal-Net \cite{Qin2019} and KaryoNet \cite{Xia2023} further enhancing robustness.
    \item \textbf{Chromosome Polarity Recognition:} Determining the correct orientation (positive/negative pole) of a chromosome in the karyogram. On the same test set, the accuracy reached 99.98\%.
    \item \textbf{Structural Aberration Detection:} This is the most challenging task, requiring the identification of subtle changes like inversions, translocations, and deletions. The HomNet model \cite{li2024chromosomal}, derived from iMedImage's multi-level features and homologous comparison concepts, achieved 95.14\% sensitivity and 100.00\% specificity in automated identification during the world's first prospective, multi-center clinical trial involving 1498 cases.
\end{itemize}

\subsection{Ultrasound Image Analysis}
iMedImage demonstrated its versatility and high performance in the ultrasound domain as well.

\subsubsection{Breast Ultrasound Mass Classification (BreastMNIST)}
We tested iMedImage on the public BreastMNIST dataset \cite{Yang2023}. This dataset contains 780 ultrasound images, with the task being to determine the presence of lesions (benign or malignant vs. normal). Using the same train/validation/test splits as the original paper, iMedImage achieved an accuracy of 96.16\%, significantly outperforming the 86.10\% accuracy reported for the Google AutoML Vision baseline model on this dataset. This highlights iMedImage's advantage in general ultrasound image classification tasks.

\subsubsection{Preterm Birth Prediction based on Cervical Ultrasound and Clinical Data}
Preterm birth (delivery < 37 weeks gestation) is a major cause of neonatal morbidity and mortality. Cervical ultrasound examination (measuring cervical length (CL), observing internal os morphology, etc.) is an important tool for predicting preterm birth risk, but its predictive power is limited and operator-dependent. Clinically, assessment usually requires combining ultrasound findings with maternal information (e.g., age, obstetric history, prior preterm birth, history of cervical surgery).
We utilized iMedImage to process multimodal data (cervical ultrasound images + maternal electronic health records) to predict preterm birth risk. The study data included 567 cervical ultrasound scans from 210 pregnant women between 16 and 28 weeks gestation, along with corresponding clinical information. Results showed that iMedImage achieved an AUC of 0.747 on the test set. For comparison, the internationally recognized QUiPP model \cite{Kuhrt2016} (developed by King's College London, based on clinical risk factors and FFN/CL measurements) achieves an AUC of approximately 0.631 on similar tasks. The performance of iMedImage significantly surpasses this benchmark, demonstrating the great potential of integrating deep features from ultrasound images with clinical information.

\subsection{CT Image Analysis}
\subsubsection{Prediction of Postoperative Recurrence in Pancreatic Cancer using Multimodal Data}
Pancreatic cancer has a high postoperative recurrence rate. Accurate prediction of recurrence risk is crucial for guiding adjuvant therapy and follow-up strategies. In collaboration with the Affiliated Hospital of Qingdao University, we used iMedImage to process multimodal data to predict 1-year Recurrence-Free Survival (RFS) risk in patients with Pancreatic Ductal Adenocarcinoma (PDAC) after surgery.
\begin{itemize}
    \item \textbf{Data:} The study integrated rich clinical and imaging data:
        \begin{itemize}
            \item \textit{Clinical Data:} Patient demographics (age, gender, BMI), preoperative laboratory tests (CA19-9, CEA, fibrinogen, etc.), detailed pathology information (tumor grade, TNM stage, lymphovascular/perineural invasion, margin status, Ki-67 index, etc.), and immunohistochemistry data (CK7/19/20, etc.).
            \item \textit{Imaging Data:} Complete preoperative contrast-enhanced CT image sequences (arterial phase, venous phase, etc.). Importantly, the model processed the raw CT images directly without requiring manual delineation of tumor regions (ROIs), achieving end-to-end feature extraction.
        \end{itemize}
    \item \textbf{Dataset:} Training set: 284 cases, Validation set: 72 cases, Test set: 89 cases. Notably, the test set data originated from an external hospital to assess model generalization.
    \item \textbf{Results:} On the external test set, the iMedImage model achieved an AUC of 0.78 for predicting 1-year RFS risk. For reference, the best-published models based on single-center data (also using multimodal data, but potentially relying on manual ROI delineation or feature extraction) report AUCs around 0.75 \cite{Liu2022}. Our results demonstrate the advantages of iMedImage in handling complex multimodal data, eliminating the need for manual annotation, and exhibiting good generalization capability.
\end{itemize}

\subsection{Generalizability Across Other Domains}
To further validate iMedImage as a general medical imaging foundation model, we tested it on a wider range of tasks involving different modalities and clinical applications. These tasks spanned from microscopic cell morphology to macroscopic organ imaging, and from common screening procedures to specialized diagnostics. Table \ref{tab:performance} summarizes the performance of iMedImage across these diverse tasks.

\begin{table}[htbp]
  \centering
  \caption{Performance of iMedImage on Various Medical Image Recognition Tasks}
  \label{tab:performance}
  \resizebox{\textwidth}{!}{
  \begin{tabular}{@{}llcccccc@{}}  
    \toprule
    \textbf{Domain} & \textbf{Task} & \textbf{Modality} & \textbf{Classes} & \textbf{Test Size} & \textbf{Accuracy (\%)} & \textbf{Sensitivity (\%)} & \textbf{Specificity (\%)} \\
    \midrule
    Cytogenetics & Chromosome Classification & Chromosome Image & 24 & 45328 & 99.96 & { - } & { - } \\
    Cytogenetics & Chromosome Polarity Recog. & Chromosome Image & 2 & 45328 & 99.98 & { - } & { - } \\
    Cytogenetics & Struc. Abnormality Recog. (Clinical Trial) & Chromosome Image & 2 & 1498 & { - } & 95.14 & 100.00 \\
    \midrule
    Cell Morphology & Bone Marrow Cell Type Recog. & Cell Image (Giemsa) & 21 & 17146 & 99.98 & { - } & { - } \\
    \midrule
    Pathology & GI Lesion Recog. (Normal/Lesion) & Pathology Slide (H\&E) & 2 & 1001 & 98.50 & 98.96 & 98.29 \\
    \midrule
    X-ray & Chest Abnormality Recog.  & Chest X-ray & 20 & 485 & 93.97 & 97.75 & 91.45 \\
    \midrule
    Endoscopy & Colorectal Lesion Recog. (Ulcer/Erosion/Polyp/Tumor) & Endoscopy & 4 & 386 & 91.66 & 92.17 & 91.39 \\
    \midrule
    Digital Camera & Neonatal Jaundice Recog. & Skin Photo & 2 & 224 & 78.57 & 84.39 & 61.11 \\
    \midrule
    Fundus Imaging & Diabetic Retinopathy Grading (G0-G4) & Retinography & 5 & 139 & 83.45 & 87.41 & 78.61 \\
    \midrule
    Dermatoscopy & Skin Lesion Recog. & Dermatoscopy & 7 & 2005 & 86.51 & 88.93 & 84.09 \\
    \midrule
    Breast Ultrasound & Breast Lesion Recog. (Malig / Benign \& Normal) & Breast Ultrasound & 2 & 156 & 88.55 & 93.27 & 86.01 \\
    \bottomrule
  \end{tabular}
  } 
  \footnotesize Note: Sensitivity and Specificity for multi-class tasks (Classes > 2) are often reported as averaged values or per-class metrics; overall values provided here might represent macro/micro averages or weighted averages depending on the specific task evaluation protocol. '-' indicates metric not applicable or not the primary focus for that task. The Breast Ultrasound task here might differ slightly from the BreastMNIST benchmark described in Sec 4.2.1.
\end{table}

As shown in Table \ref{tab:performance}, iMedImage effectively handles diverse modalities ranging from microscopic (chromosomes, cells) to macroscopic (chest X-rays, CT), from 2D (X-ray, pathology slides) to 3D (CT), and utilizing different imaging principles (light microscopy, ultrasound, X-ray, standard camera). Across these tasks, iMedImage consistently demonstrated high performance, matching or exceeding existing specialized models in several domains. For example, in the task of classifying 7 types of skin lesions including melanoma using dermatoscopy images, the model achieved an average accuracy of 86.51\%, significantly higher than previously reported state-of-the-art levels (approx. 76.8\%).

These broad successful applications validate iMedImage's high baseline performance, low data dependency (achieved through pre-training), and low-cost transferability as a foundation model. The model's strong generalization capability stems from its unified architecture design and pre-training on large-scale, diverse medical image data.

Furthermore, the iMedImage model has completed the Internet Information Service Algorithm Filing in China (Record Number: 330113553162801240011) and passed relevant model evaluation tests conducted by the China Academy of Information and Communications Technology (CAICT), laying a regulatory and technical foundation for its subsequent clinical translation and industrial application.

\section{Discussion}
This study successfully developed and validated iMedImage, a foundation model oriented towards general medical image recognition. Its core advantage lies in effectively integrating information from diverse modalities and clinical domains through a unified architecture and pre-training strategy, learning highly generalizable visual feature representations. Experimental results demonstrate that iMedImage not only achieves breakthrough progress in its core application—chromosome abnormality detection (particularly in structural abnormality recognition)—but also exhibits strong performance and broad applicability across multiple other domains, including cytology, pathology, radiology (X-ray, CT), ultrasound, endoscopy, fundus imaging, dermatoscopy, and even standard digital photography.

In the most challenging task of chromosome abnormality detection, iMedImage achieved significant breakthroughs. Especially in structural abnormality recognition, derivative models like HomNet \cite{li2024chromosomal}, incorporating multi-level analysis and modeling of homologous chromosome differences, enabled high-accuracy, fully automated detection for the first time, validated in a large-scale prospective clinical trial. This holds substantial significance for improving the efficiency of genetic disease diagnosis, reducing the burden on technicians, standardizing analysis workflows, and potentially contributing significantly to reducing birth defects caused by chromosomal abnormalities. The use of a multi-center, multi-device test set with naturally distributed data further substantiates the model's robustness and clinical utility.

The generalizability of iMedImage was confirmed in other domains like ultrasound and CT. Whether processing 2D ultrasound images for mass classification, fusing ultrasound images with clinical text data for preterm birth prediction, or analyzing 3D CT images combined with complex clinical-pathological data for pancreatic cancer recurrence prediction, iMedImage demonstrated performance superior to existing benchmarks or specialized models. Particularly noteworthy is its ability to process raw CT images end-to-end without manual ROI annotation in the pancreatic cancer prediction task \cite{Liu2022}, achieving excellent performance on an external test set, which significantly enhances the model's usability and generalization potential in real-world clinical workflows.

The key techniques employed by iMedImage, such as multi-level analysis, CoT embedding \cite{Wei2022}, and MoE strategy \cite{Shazeer2017}, played crucial roles in its success. Multi-level analysis allows the model to concurrently focus on details (e.g., chromosome bands, lesion textures) and global context (e.g., organ structure, overall case presentation). CoT embedding aids the model in tackling tasks requiring complex logical reasoning. MoE enhances the model's flexibility and efficiency in handling heterogeneous data.

This study, however, has limitations. Although we constructed a large-scale multimodal dataset, there is still room for improvement in data diversity (e.g., rare diseases, different ethnic groups, more scanner types). For certain applications (e.g., preterm birth prediction, pancreatic cancer recurrence prediction), while achieving superior performance compared to benchmarks \cite{Kuhrt2016, Liu2022}, larger-scale, long-term prospective studies are needed to fully validate their clinical utility. Furthermore, although strategies like CoT can offer some insight into the model's decision process, the interpretability of deep learning models remains an ongoing challenge, especially in high-stakes medical decision-making scenarios.

Future work will focus on the following aspects:
\begin{itemize}
    \item \textbf{Continuous Expansion of Pre-training Dataset:} Incorporating more modalities (e.g., PET, ophthalmology images), more disease types, and data from broader geographical and demographic sources.
    \item \textbf{Model Optimization and Extension:} Further research into more efficient and interpretable model architectures; enhancing capabilities for processing 3D and 4D (time-series) images; exploring deeper fusion with other modalities like electronic health records (text) and genomics data.
    \item \textbf{Downstream Task Expansion:} Applying iMedImage to more clinically valuable medical image analysis tasks.
    \item \textbf{Clinical Validation and Deployment:} Conducting larger-scale prospective clinical trials for promising applications (e.g., chromosome abnormality detection, cancer risk prediction) and exploring feasible solutions for integrating the model into clinical workflows, such as via cloud platforms or localized deployment.
    \item \textbf{Federated Learning and Privacy Preservation:} Investigating federated learning frameworks to train and optimize models using multi-center data while preserving data privacy.
\end{itemize}

\section{Conclusion}
We presented iMedImage, an innovative, end-to-end general foundation model for medical image recognition. Through unified representation, a unified model architecture, and unified cross-domain pre-training, iMedImage effectively handles multiple medical image modalities and tasks spanning cytogenetics, pathology, radiology, ultrasound, and more. We highlighted its breakthrough performance in automated chromosome karyotype analysis, particularly in detecting structural abnormalities, validating its high sensitivity and specificity on large-scale multi-center clinical data. Furthermore, the successful application of iMedImage across various other clinical tasks demonstrates its broad generalizability, high baseline performance, and potential for low-cost transferability. iMedImage offers a powerful auxiliary tool for clinicians, promising to significantly enhance the efficiency and accuracy of medical image analysis, ultimately contributing to improved patient outcomes and reduced healthcare burdens.


\end{document}